# HR-CAM: Precise Localization of pathology using multi-level learning in CNNs

Sumeet Shinde[1*], Tanay Chougule[1*], Jitender Saini[2] and Madhura Ingalhalikar[1]

[1] Symbiosis Center for Medical Image Analysis, Symbiosis International University, Pune 412115, India
`head@scmia.edu.in`
[2] National Institute of Mental Health and Neurosciences, Bengaluru 560029, India

**Abstract.** We propose a CNN based technique that aggregates feature maps from its multiple layers that can localize abnormalities with greater details as well as predict pathology under consideration. Existing class activation mapping (CAM) techniques extract feature maps from either the final layer or a single intermediate layer to create the discriminative maps and then interpolate to upsample to the original image resolution. In this case, the subject specific localization is coarse and is unable to capture subtle abnormalities. To mitigate this, our method builds a novel CNN based discriminative localization model that we call high resolution CAM (HR-CAM), which accounts for layers from each resolution, therefore facilitating a comprehensive map that can delineate the pathology for each subject by combining low-level, intermediate as well as high-level features from the CNN. Moreover, our model directly provides the discriminative map in the resolution of the original image facilitating finer delineation of abnormalities. We demonstrate the working of our model on a simulated abnormalities data where we illustrate how the model captures finer details in the final discriminative maps as compared to current techniques. We then apply this technique: (1) to classify ependymomas from grade IV glioblastoma on T1-weighted contrast enhanced (T1-CE) MRI and (2) to predict Parkinson's disease from neuromelanin sensitive MRI. In all these cases we demonstrate that our model not only predicts pathologies with high accuracies, but also creates clinically interpretable subject specific high resolution discriminative localizations. Overall, the technique can be generalized to any CNN and carries high relevance in a clinical setting.

## 1    Introduction

Convolutional neural networks (CNNs) have become one of the most powerful tools for analyzing medical images and have demonstrated exceptional performance in image classification tasks [1] as well as in structure segmentation [2] often becoming the new state of art in several use cases. The ability of CNNs in extracting the discriminative pixel based features, automatically, is attractive as it often accounts for bet-ter classification performance, particularly on large datasets, when compared to a similar task on empirically drawn features. This superior performance, however, comes with certain tradeoffs as the predictions are not intuitive and explanation of objects/regions that contribute towards the classification is crucial, especially for problems in medical imaging where gaining clarity is particularly imperative as the classification output has to be supported with clinical interpretation.

---

*both authors have equal contribution.

To this date, multiple techniques have been proposed to visualize the discriminative regions that allow us to gain insights into the functioning of the CNNs. Earlier, techniques were based on mapping activations back to the input space via deconvolutions to highlight the most discriminative regions [3]. However, the applicability of this deconv-net was limited owing to its complexity. Existing techniques are mainly based on class activation mapping (CAM) that have the capability to localize the regions either by extracting information from the final layer[4] using global average pooling (GAP) or any layer of choice using gradient information (Grad-CAM)[5]. These techniques are more applicable, and further have been adapted for medical imaging 2D and 3D applications [6, 7], for extracting multiple object instances [8] and for refining the classification output by revisiting the discriminative regions [9].

The challenge with CAMs is that they provide insights usually only from a single layer of the CNN. Moreover, the extracted CAM is a low resolution map especially when extracted from intermediate to final layers of CNN as the input is max-pooled multiple times. This facilitates a coarse output which fails to capture subtle details that discriminate the classes under consideration. Particularly, this is crucial in medical imaging, where the pathology under consideration may demonstrate highly heterogeneous and diffuse patterns of abnormalities that may not be characterized by current CAM techniques. It is therefore important to extract these subtle discriminative image based markers as these can support prognosis, diagnosis, risk prediction and treatment efficacy.

In this paper we address the aforementioned issues. We create high resolution class activation maps (HR-CAMs) to visualize and highlight the discriminative regions in pathologies with high precision. We achieve this by capturing feature maps from multiple layers (pre-max-pooling layers) and aggregating these for a single layer classifier that also facilitates improved classification accuracies. We first validate our novel technique on simulated abnormalities dataset where we demonstrate its superior working even on noisy data and then apply it to (1) predict Parkinson's disease from neuromelanin contrast MRI (brain-stem region) (2) classify ependymomas from glioblastomas on T1-CE MRI. The multi-layer aggregation provides better interpretability with higher accuracy while the HR maps facilitate precise localization. The method gives state-of-art performance when compared to current techniques, and is generalizable to any CNN model.

## 2 Method

In this work, we demonstrate that by employing a novel CNN based architecture, we can automate and generalize its ability in identifying the exact regions of an image that are important in the classification by generating HR-CAMs. Figure 1 gives the overview of our HR-CAM architecture.

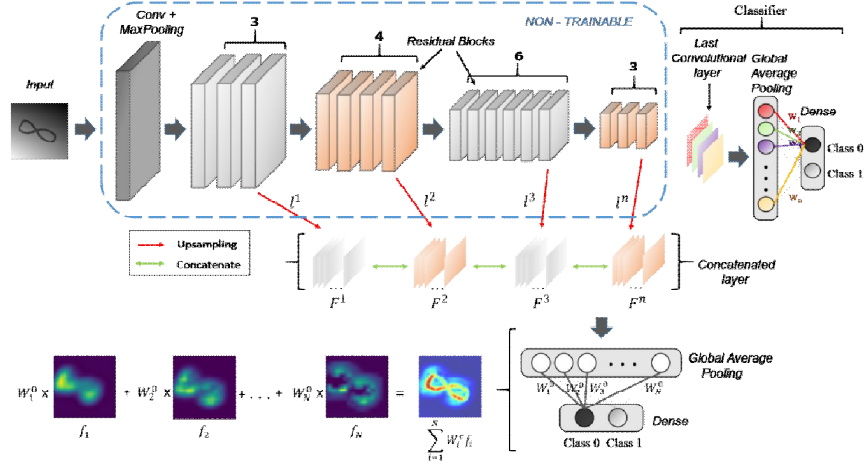

**Fig. 1:** Schematic diagram of the proposed model. The input images are first trained using CNN layers with a classifier at the end. The feature maps of each layer before max-pooling are sampled and concatenated and given as input to a global average pooling layer that is trained to optimize the weights $(W_1, W_2 ..., W_N)$ while keeping the previous layers frozen. These weights are used to create the HR-CAMs.

### 2.1 HR-CAM

Specifically, if we choose $k$ number of convolutional layers $L=\{l^1,l^2,...,l^k\}$, the feature maps $F^i=\{f_1^i, f_2^i..., f_m^i\}$ corresponding to the convolutional layers $l^i$ are first upsampled using a bilinear interpolation to match the original input image dimensions. The number of feature maps corresponding to the $k^{th}$ layer is given by $m_k$. The feature maps are then stacked together to form a concatenated layer with total $N$ feature maps where $N$ is given by equation 1.

$$N = \sum_{i=1}^{k} m_k \tag{1}$$

These $N$ feature maps are then average pooled globally thereby representing every feature map in the concatenated layer by a single value. The vector obtained after average pooling is given as input to a single fully-connected layer that is trained to minimize the categorical cross entropy loss function (equation 2), where $y_n$ is target output probability, $\hat{y}_n$ is predicted output probability, $S$ is number of samples and $J(w)$ is categorical cross-entropy loss and is achieved by using the Adam optimizer. The optimization is performed by freezing the convolutional layers.

$$J(w) = \frac{-1}{S}\sum_{n=1}^{S}[\,y_n\,log\,\hat{y}_n\, +\, (1-y_n)\,log(1-\hat{y}_n)] \qquad (2)$$

This assigns new weights to all the feature maps according to their importance for classification and supports weighted aggregation. Finally, to obtain the discriminative activations, we forward propagate the input image and acquire the weights $(W_1, W_2 ..., W_N)$ at the output layer for the respective class, as given in Zhou et al. [4]. To create the class activation map, the weights $W_i^c$ for the respective class are then multiplied with the feature maps $f_i$ and then added together, as shown in Figure 1. The resulting class activation map $A$ can therefore demonstrate the most discerning regions in the image in high resolution.

$$A = \sum_{i=1}^{N} W_i^c f_i \qquad (3)$$

Although figure 1 uses Resnet50 architecture [10] with global average pooling (GAP) and fully connected layers for binary classification, the proposed technique is generalizable to any CNN based classifier.

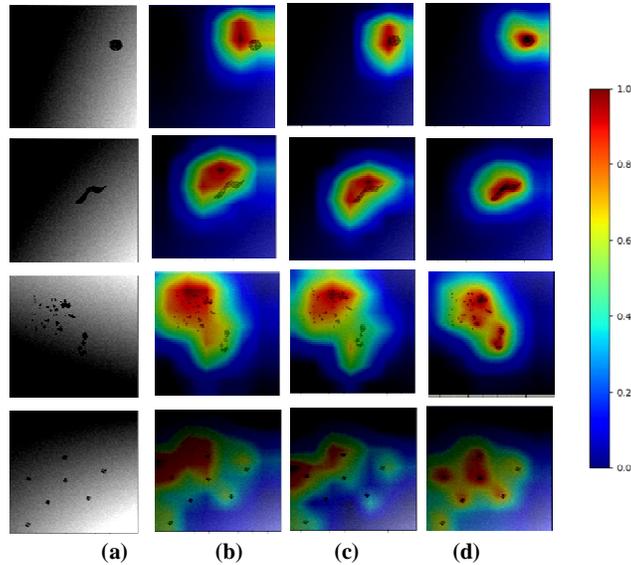

**(a)  (b)  (c)  (d)**

**Fig. 2:** Simulated dataset results. (a) demonstrates the simulated abnormities, (b) CAMs produced using Zhou et al.'s method (c) results from GradCAM (d) results from HR-CAM (proposed technique). The maps produced from our method are high resolution as can be clearly seen (especially in top 2 rows). Moreover, in case of diffuse abnormalities, our method can capture it more finely (bottom 2 rows).

## 3  Experiments and Results

### 3.1  Datasets

Our dataset consisted of two classes of simulated images. Class 1 included 1000 images of normative data without abnormalities but with added random noise and Class 2 consisted of 1000 images with simulated pathology with added random noise. We created abnormalities that were localized as well as abnormalities that were diffuse in nature as shown in Fig. 2. The data was divided into train and test groups with 1500 images for training and 500 for testing.

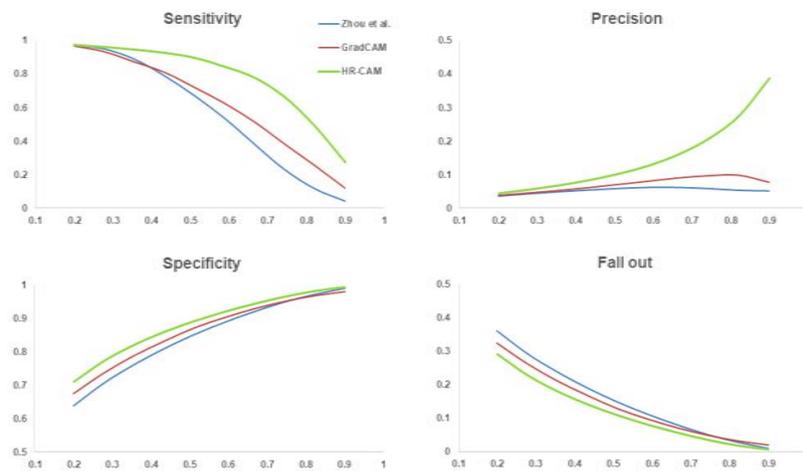

**Fig.3:** Comparative quantitative analysis of the three CAM methods. The x-axis in all 3 plots denotes the threshold level for the CAMs. We demonstrate that at any threshold our HR-CAMs are more sensitive, specific with a very low fall out and high precision when the simulated abnormalities were considered as ground truth.

The second dataset included high resolution 3D neuromelanin sensitive MRI (NMS-MRI) scanned using spectral pre-saturation with inversion recovery (SPIR) sequence and was acquired using TR/TE: 26/2.2ms, flip angle: 20º; reconstructed matrix size: 512 x 512; field of view: 180x180x50mm, for 45 patients with Parkinson's disease (PD) and 35 controls. The NMS-MRI provides a good contrast in the substantia nigra pars compacta (SNc) and is useful in identifying PD as PD manifests depigmentation of the SNc[11]. The CNN was employed on the boxed region around the brainstem on the axial slices, with 30 PD and 25 controls in training and the remaining for testing.

Our final dataset included T1 contrast enhanced (T1-CE) MRI for 26 cases of ependymomas and 26 cases of grade IV glioblastoma. Majority ependymoma's tend to mimic grade IV glioblastomas and therefore it is crucial to be identified at a radio-logical level. We used a large boxed region around the lesion in the axial slices as an input to the CNNs with 3811 images for training with augmentation and 1004 images for testing.

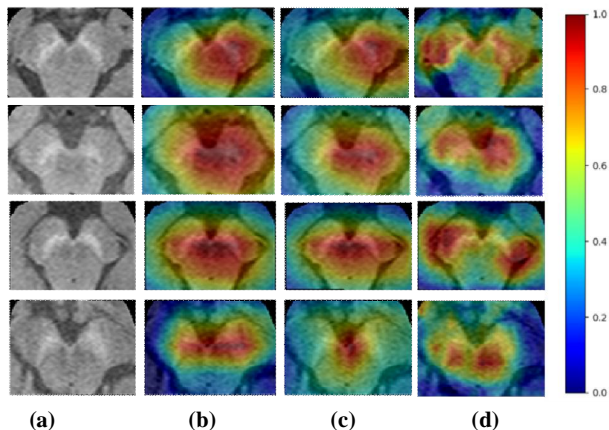

(a)       (b)       (c)       (d)

**Fig 4:** Qualitative Comparison of Class Activation Maps obtained on the neuromelanin contrast images to predict PD. (a) Section of brainstem used in classification where the bright region is the SNc. (b) Results from Zhou.et al.'s CAMs (c) Results from GradCAM (d) Results from our proposed technique (HR-CAMs). Our methods can delineate the abnormal areas (SNc in this case) better than both the comparative techniques that produce a diffuse map.

### 3.2 Experiment details

The CNN architecture was derived from ResNet50 design that consisted of 16 residual blocks each of which contains 3 convolutional layers and an identity connection. A max-pooling layer, that downsamples the image, was applied before the first block and after 3, 4, 6 and 3 residual blocks as shown in Figure 1. The final layers included a GAP layer followed by fully connected layers and a 'softmax' classifier. Throughout the CNN architecture, we employed rectified linear unit (ReLU) activations and a learning rate of 0.0001. The cross entropy loss was minimized as shown in equation 2. To reduce the susceptibility to over-fitting, as a standard practice, data augmentation was performed which increases the dataset by several folds. We performed image translations, rotations, minor shifting as well as horizontal and vertical flipping. To increase the total dataset, we applied a random combination of these transformations on each image.

To compare our results with existing techniques we employed GAP based CAMs by Zhou et al [4] that uses the final layer feature maps and Grad-CAMs where we used the layer before the final layer on the same ResNet50 model. Our models were trained on Nvidia Quadro P6000 with 24GB of graphics RAM. The code was implemented in Python 3.6 and the deep-learning Framework used was Keras with a TensorFlow backend.

A discriminative ground truth area is necessary for quantifying the localization ability of CAMs. Therefore, we compared the three techniques- Zhou-CAMs, Grad-CAM, and HR-CAM (ours), with simulated abnormalities as the ground truth. All the class activation maps were intensity normalized (0 to 1.0) and compared with the ground truth masks ($A_T$) by varying threshold from 0.1 to 0.9. For every threshold, a binary image ($A_C$) was obtained and the foreground pixels were considered as positive labels ($P_c$) and the rest as negative labels ($N_c$). These were then matched with the ground-truth positive ($P_T$) and negative ($N_T$) labels where $P_T$ is given by the abnormality and $N_T$ is the noisy background. We compared the CAM localization performance based on sensitivity, specificity, precision and fall-out.

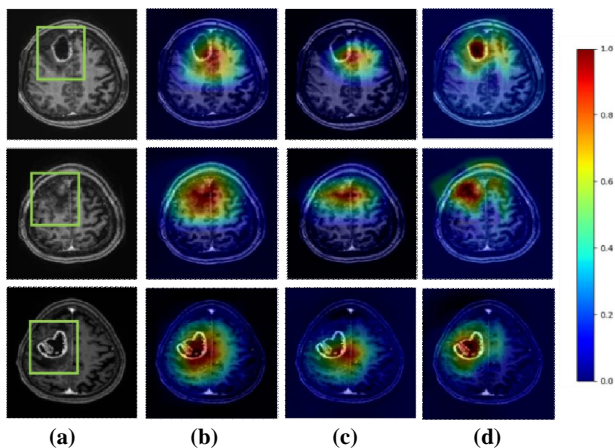

(a)      (b)      (c)      (d)

**Fig 5:** Figure showing results for tumor classification (a) original image (boxed region is used for training/testing) (b) Results from Zhou et al.'s CAMs (c) Results from GradCAM (using one layer before the final layer) (d) HR-CAMs (proposed technique). It can be clearly seen by optimizing the weights and summing the feature maps from multiple layers, we obtain more accuracy in locating the abnormality.

### 3.3 Results

Table 1: Quantitative comparison of CAMs on simulated data.

| Mean values | Zhou et. al. | GradCAM | HR-CAM (proposed) |
|---|---|---|---|
| Sensitivity | 0.556 | 0.620 | **0.774** |
| Specificity | 0.848 | 0.863 | **0.885** |
| Precision | 0.052 | 0.072 | **0.151** |
| Fall out | 0.151 | 0.136 | **0.114** |

Fig. 2 shows the comparison of the CAMs between our HR-CAM, Zhou's CAMs and Grad-CAM for the abnormal test cases. Table 1 and Fig. 3 provide the comparative quantitative analysis for the simulated CAMs. Table 1 provides the mean values for the evaluation quantifiers over all thresholds. It can be observed that HR-CAMs can accurately visualize the abnormalities than earlier techniques. For the PD dataset the HR-CAM model boosts the accuracy to 78.3% when compared to 76.2% with the ResNet50. Fig. 4 demonstrates the heatmaps for PD test cases in comparison to Zhou's CAMs and Grad-CAM wherein it can be clearly seen that our method captures the SNc area (right and left separately) while the other two provide a very coarse activation map with the left-right SNc merged in a single activation blob. For the tumor data we demonstrate similar findings where the classification accuracy is boosted to 67.5% when compared to ResNet50 – 64.6%. Fig. 5 demonstrates the CAMs computed for the tumor test cases. By employing the HR-CAM technique, we achieve higher classification accuracy as well as a detailed heatmap.

## 3      Conclusion

This work presented a novel CNN based technique named high resolution class activation mapping (HR-CAMs) that can generate precise maps of the most discriminative areas involved in distinguishing the groups under consideration. We demonstrated the accuracy of capturing the abnormalities through simulated data as well as through multiple applications. In summary, the presented framework is highly relevant to classification problems in medical imaging where the identification of discriminative regions is crucial for clinical rationalization as well as may aid in targeted therapy and treatment planning.